# Advanced Synthetic Image Detection Using Hybrid Feature Extraction.

[1]S. Navaneetha Krishnan, [2]Ameya Uppina, [3]Nikhil Iyer, [4]Joe Dhanith P R

[1,2,3,4]School of Computer Science and Engineering, Vellore Institute of Technology Chennai Campus, India

Email: [1]krishnannavaneetha467@gmail.com, [2]ameyauppina2003@gmail.com, [3]nikhiliyer95@gmail.com, [4]joedhanith.pr@vit.ac.in

**Abstract:** In the realm of generative AI, models such as GANs, MidJourney, DALL-E, and others have been continuously evolving to generate increasingly lifelike images. This rapid advancement has created an urgent need for robust methods to distinguish between real and AI-generated images. This paper proposes a novel approach for the detection of synthetic images by employing deep learning techniques for feature analysis. The proposed architecture classifies images by identifying variations that generative models struggle to replicate, particularly fine-grained textures and complex object features. Deformations in complexes are examined with the help of the CBAM architecture along with ResNet50. We have performed a selective patch-based analysis using Discrete Cosine Transform (DCT) scoring, extracting the regions with the most and least complexity for further processing with an SRM Module and ResNet50 module. Experimental results show that our approach achieves high accuracy in classifying real-world images from synthetic ones, contributing to the development of more reliable detection frameworks.

## I. INTRODUCTION

With the rise of Artificial Intelligence, it has become harder to distinguish between real and artificially generated images. The rise of Generative Models has enabled the creation of very realistic images. These advancements might be very helpful in a wide range of applications, such as art generation and data augmentation, but they pose a potential risk of spreading misinformation and potential identity fraud.

Primarily, Image generation is done using three types of image generation models, namely the Generative Adversarial Network, Diffusion Models, and Midjourney.

The rise of Generative Adversarial Networks (GAN) [4], which is a class of machine learning models intended to generate highly realistic synthetic images, videos, and text. The GAN has two neural networks: the generator and the discriminator. The generator's goal is to create images that closely resemble real ones, while the discriminator's job is to distinguish between real and generated images. The generator continuously improves through repeated training, learning to produce images that become increasingly difficult for the discriminator to detect as fake. Thus, making them capable of generating more realistic looking images makes it a powerful tool in the generation of deepfakes and Image enhancements.

Diffusion Models are a type of generative model that create images by starting with a random noise image and gradually refining it by a step-by-step de-noising process. A diffusion model generates images through two key steps: forward diffusion and reverse diffusion. In forward diffusion, noise is gradually added to an image during training until it becomes completely unrecognizable. During reverse diffusion, the model starts with pure noise and systematically removes it step by step, using a neural network to predict and eliminate noise at each stage, ultimately reconstructing a realistic image. To distinguish images generated by such models from real images, several methods are currently being explored.

Midjourney's architecture combines natural language processing and image generation, utilizing CLIP for text interpretation and iterative refinement for enhanced image quality. A text encoder and generative model ensure semantic alignment between prompts and generated images.

Vision Transformers (ViT) [25] process specific image patches to show a high accuracy in detecting subtle inconsistencies With these methodologies in mind, we propose a novel feature-based extraction technique to solve this problem.

## II. LITERATURE SURVEY

The paper [13] proposed by Despina K et al. explores improving synthetic image detection using texture-based cropping techniques. The study utilizes high-resolution images from datasets such as Forensynths, Synthbuster, TWIGMA, RAISE, and OpenImages. The datasets include images generated by models like StyleGAN, DALL-E 2, and SD 1.5, alongside real images from OpenImages. The study introduces a novel method called TextureCrop, which involves cropping images based on texture features to enhance the detection of synthetic images. This approach focuses on the idea that synthetic images often lack the rich texture diversity present in real images. By cropping images to highlight these texture differences, the model can better distinguish between real and synthetic images. The authors evaluate the effectiveness of TextureCrop by integrating it with existing detection models and comparing the results with baseline methods. The proposed TextureCrop method significantly enhances detection performance compared to baseline methods. It shows an average increase of 12.1% in Accuracy , 18.2% in Precision , and 14.9% in AUC compared to resizing. Compared to CenterCrop, TextureCrop offers a 4.3% increase in Accuracy, a 6.9% increase in Precision, and a 6.1% increase in AUC. Among the methods tested, PatchCraft demonstrates the largest gains with TextureCrop, showing a 10.00% increase in Accuracy, a 9.14% increase in Precision, and a 9.64% increase in AUC.

The Architecture [14] proposed by U Ojha et al. aims to develop a universal fake image detection framework that can generalize across various generative models. This involves designing a detection framework that can effectively capture the artifacts and patterns common to AI-generated images, regardless of the specific generative model used. The architecture employs a combination of feature extraction techniques and robust training strategies to ensures the model's ability to generalize well. The study involves training and testing on datasets with images generated by multiple models, primarily focusing on 20 different ProGANs trained on various domains. The impact of dataset diversity is explored by considering datasets with 2, 4, 8, and 20 classes, as well as combining ProGAN and LDM images from two different domains. The proposed approach shows significant improvements over baseline methods. The nearest neighbor and linear probing variants achieve high accuracy and average precision across different generative models, including GANs and diffusion models. For instance, the nearest neighbor variant achieves an average accuracy of 93% on GANs and 82-84% on diffusion models, outperforming trained deep networks by 8-30%. Notably, the approach maintains high performance on unseen models like DALL-E, with improvements of +25-30% in average accuracy. This study highlights the effectiveness of simple yet robust methods in detecting fake images across diverse generative models, often surpassing the performance of complex deep learning networks.

This study [10] by N Zhong et al. introduces a novel preprocessing technique called Smash & Reconstruction, which erases global semantic information and enhances texture patches. This approach involves dividing an image into patches and reconstructing them into two separate images: one for rich-texture patches and another for poor texture patches. The authors then extract the inter-pixel correlation discrepancy feature between these regions, which serves as a universal fingerprint for AI generated image detection across different models. This feature is particularly effective because synthesizing realistic rich texture regions is more challenging for existing generative models. The study involves training and testing on datasets containing images generated by 16 different models, including GAN-based models and DALL-E 2. The training set consists of 720,000 images: 360,000 real images from LSUN and 360,000 fake images from ProGAN. This diverse dataset allows for a thorough evaluation of the method's generalization capabilities, especially since the detector does not know the generative model of the fake images during practical applications. The proposed methodology outperforms existing detectors by a substantial margin, achieving satisfactory results for most generative models. However, some performance degradation is noted for images generated by CycleGAN and GauGAN due to their distinct generation process. The average detection accuracy of the proposed method is 89.31%. When evaluating robustness to image distortions, the method maintains superior performance compared to baselines, with detection accuracies of 72.48% for JPEG compression, 78.36% for downsampling, and 75.99% for Gaussian blur. This indicates the proposed method's robustness and effectiveness in real-world scenarios where images may be distorted.

The paper [8] by Hina Otake et al. explores using synthetic data-driven representations to detect synthetic images, a critical task due to the misuse of AI-generated images. The study investigates the effectiveness of synthetic data-driven representations for Synthetic Image Detection by leveraging vision transformers trained with the latest visual representation learners. Specifically, it uses StableRep and SynCLR, which are trained exclusively on synthetic data generated by diffusion models. The UnivFD framework is employed as the synthetic image detector, with different pre-trained foundation models serving as the backbone. The authors compare the performance of these models against widely used baselines like CLIP and DINOv2, which are trained on real data. Additionally, an ensemble learning approach is explored to combine the features of models trained on synthetic and real data. The study evaluates detection performance across various generative models, including GAN-based methods and diffusion model-based methods , as well as other methods like DeepFakes and DALL-E. The datasets used include ProGAN's training data for training a fully connected layer in UnivFD. The study reports significant improvements when using synthetic data-driven representations. Specifically, SynCLR shows a performance improvement of +10.32 mAP and +4.73% accuracy over CLIP when detecting images from GANs not used during pre-training. The ensemble of CLIP and SynCLR further enhances performance, with improvements of +7.53 mAP and +0.58% accuracy compared to the baseline. However, the detection performance for images generated by diffusion models remains relatively low, indicating a need for further refinement. This suggests that while the approach is effective for GAN-generated images, additional strategies are required to improve detection capabilities for diffusion models.

The paper [12] by Z Xi et al, involves employing a dual-stream network architecture, where one stream processes local features and the other stream processes global features. The cross-attention mechanism is employed to enhance the interaction between these two streams, allowing the model to capture both fine-grained details and overall semantic information. This approach enables the model to identify subtle differences between real and AI-generated images more effectively. It involves training and testing on two text-to-image (T2I) databases, DALL·E2 and DreamStudio, each containing 20,000 real and AI-generated images. The images are resized to three resolutions (256×256, 128×128, and 64×64) and subjected to JPEG compression to simulate real-world scenarios. The model is also evaluated on the SPL2018 and DsTok datasets for computer-generated (CG) detection. The proposed approach demonstrates significant improvements in detection performance compared to baseline methods. On the DALL·E2 and DreamStudio datasets, the model achieves high accuracy, with detection rates exceeding 98.5% at a resolution of 256×256. Even at lower resolutions (64×64), the model maintains impressive performance, achieving 93.1% and 97.8% accuracy for the two T2I databases, respectively. The model shows robustness against various post-processing operations, outperforming other methods in terms of average accuracy. This indicates that the cross-attention enhanced dual-stream network is effective in detecting AI-generated images across different resolutions and post-processing conditions, making it suitable for real-world applications where image quality may vary.

The study [9] by FM Rodríguez et al, focuses on detecting photorealistic AI-created images using pixel-wise feature extraction techniques combined with CNNs. The study employs two primary pixel-wise feature extraction techniques: PRNU and ELA. PRNU (Photo Response Non-Uniformity) exploits the unique noise patterns present in images captured by physical cameras, which are absent in AI generated images. Error level analysis (ELA), traditionally used for detecting image editing, analyzes the compression artifacts in images. These features are then used to train CNNs, which are effective in distinguishing visually different classes. The CNNs learn to differentiate between real and AI-generated images based on the extracted features, allowing for a binary classification decision. The dataset includes images created with DALL-E, Stable Diffusion, and OpenArt, alongside real camera photographs from databases like the Dresden Image Database and the VISION dataset. The Photo Response Non-Uniformity (PRNU) method achieves an accuracy of 0.95, with a precision of 0.93, recall of 0.97, and an F1-score of 0.95. The Error Level Analysis (ELA) method performs even better, with an accuracy of 0.98, a precision of 0.97, recall of 0.99, and an F1-score of 0.98. These metrics indicate strong performance in detecting AI-generated images using both PRNU and ELA features. The study highlights the effectiveness of combining pixel-wise feature extraction with CNNs for distinguishing between real and AI-generated images, achieving high accuracy rates that are crucial for image forensics applications.

The paper [16] by Mert Bulent Sariyildiz et al. explores leveraging synthetic data to enhance the transferability of representations in deep learning models. The study involves creating synthetic clones of the ImageNet dataset using generative models. These clones are designed to mimic the structure and diversity of the original ImageNet dataset but are generated synthetically. The authors employ a combination of feature learning techniques and robust training strategies to ensure that the models can generalize well to unseen data. They also explore the impact of different generative models on the quality and transferability of the learned representations. The study involves training on synthetic ImageNet clones generated by Stable Diffusion, referred to as ImageNet-SD. These datasets are created by synthesizing images for each ImageNet class using class names as prompts, augmented with hypernyms and definitions from WordNet, and diverse backgrounds from the Places dataset. On ImageNet-100, a guidance scale of 2 achieves the best performance with over 63% Top-1 accuracy, further improving to 73.3% when scaling synthetic images 50×50 . On ImageNet-1K, the model trained on synthetic data attains 42.9% Top-1 accuracy, demonstrating its ability to capture fine-grained distinctions despite trailing real-image-trained models. In transfer learning, synthetic data representations generalize well, reaching 70.4% Top-1 accuracy on ImageNet-1K (CoG benchmark), highlighting its viability when real data is scarce.

The paper [15] by D. Gragnaniello et al. critically analyzes the detection of GAN-generated images. The study involves training and testing on datasets that include images generated by multiple GAN models. The study involves evaluating several state-of-the-art detection methods using a variety of GAN models. The authors employ a combination of feature extraction techniques and robust training strategies to ensure that the models can generalize well to unseen data. They also explore the impact of different GAN architectures on detection performance, highlighting the challenges in maintaining robustness across diverse image types. The training dataset consists of 362,000 real images from the LSUN dataset and 362,000 generated images from 20 different ProGAN models. Testing datasets include low-resolution (256x256) and high-resolution (1024x1024) images from various GAN architectures. Real images are sourced from ImageNet, COCO, and RAISE datasets. While the AUC is generally high for low-resolution images, accuracy at a fixed threshold and probability of detection at low false alarm rates (FAR) are less encouraging. For high-resolution images, performance is somewhat better but still faces challenges with compression and resizing. The best variant, which avoids down-sampling in the first block, achieves an average gain of about 15% in accuracy and 14% in Pd 5% compared to the baseline. Training on StyleGAN2 models further enhances performance, with accuracy often exceeding 97%. This highlights the effectiveness of tailored approaches in improving the detection of GAN-generated images, particularly when optimized for specific architectures like StyleGAN2.

The paper [11] by Edmar R. S. de Rezende et al. focuses on detecting computer-generated images using deep convolutional neural networks (CNNs) and transfer learning techniques. The study employs Residual Networks (ResNets) as the base architecture for developing different models adapted to the binary classification problem of identifying computer-generated images. Unlike traditional approaches, these models are trained on raw pixel data without relying on hand-crafted features. This approach allows for more stable results with lower variance compared to state-of the-art methods. The authors utilize transfer learning to adapt the models to the specific task of distinguishing between real and computer-generated images. The study uses the DSTok and DSTok-Ext datasets for evaluation. DSTok consists of 4850 computer-generated images (CGIs) and 4850 natural images (NIs), featuring diverse indoor and outdoor landscapes with photorealistic CGIs. DSTok-Ext extends this collection, providing additional images for comprehensive testing. The proposed method achieves a detection accuracy comparable to state-of-the-art methods, reaching an accuracy of 0.97. This performance is notable because it is achieved without manual feature extraction, a significant advantage over traditional methods. The stability of the results, indicated by lower variance, suggests that the model is robust across different image types. The use of these datasets allows for a thorough assessment of the model's ability to generalize across different image types and generative models. The high-resolution images and significant inter-class diversity in the datasets make them pivotal for research in CG image detection. Overall, the study demonstrates the effectiveness of deep learning in detecting computer-generated images, highlighting the potential of CNNs to handle complex image classification tasks without extensive manual intervention.

In the paper [26] proposed by S Yan, they went through several studies to explore AI-generated image detection using diverse architectures, feature extraction techniques, and datasets to enhance detection robustness. Early CNN-based approaches, such as CNNSpot and GramNet, focus on spatial features but struggle with generalization, while transformer-based models like Swin-T and DeiT-S improve contextual understanding but suffer performance degradation on unseen generative models. AIDE (AI-generated Image Detector with Hybrid Features) integrates patchwise low-level statistical analysis and high-level semantic embedding for robust AI-generated image detection. The Patchwise Feature Extraction (PFE) module utilizes Discrete Cosine Transform (DCT) to score patches based on frequency content, identifying high- and low-frequency regions. These patches undergo Spatial Rich Model (SRM) filtering to extract noise patterns and are processed by ResNet-50 to obtain frequency-based feature embeddings. The Semantic Feature Embedding (SFE) module leverages the OpenCLIP ConvNeXt model to capture object co-occurrence and contextual relationships, followed by a linear projection and mean spatial pooling for feature aggregation. The Discriminator fuses frequency-based and semantic features through channel-wise concatenation and an MLP classifier to distinguish AI-generated images from real ones. This hybrid approach ensures improved robustness against diverse generative models, effectively detecting subtle artifacts while preserving high-level visual coherence. These models leverage datasets like AIGCDetectBenchmark, GenImage, and Chameleon, encompassing AI-generated images from StyleGAN, BigGAN, MidJourney, and Stable Diffusion v1.5, ensuring diverse training data. AIDE achieves 92.77% accuracy across multiple generative models, outperforming prior methods by 3.5% in AUC and 4.6% in average precision, using classification accuracy, average precision (AP), and AUC as key evaluation metrics. Despite these improvements, challenges persist due to evolving generative techniques, necessitating further research into adversarial robustness, dataset diversity, and multimodal fusion for more reliable AI-generated image detection.

## III. DATASETS

The advancement of AI-generated images has been largely driven by powerful image-generation models such as Stable Diffusion v1.5, GANs, MidJourney, DALL-E, and Imagen3. Each of these models employs different techniques to generate realistic or artistic images. Stable Diffusion v1.5 is a diffusion-based generative model known for its ability to create highly detailed images with user-defined prompts. It operates by iteratively denoising a latent representation of an image until the final version emerges. Generative Adversarial Networks (GANs) consist of two competing neural networks—the generator and the discriminator—that work together to produce highly realistic images. The generator creates images, while the discriminator evaluates their authenticity against real images. Through this adversarial process, GANs have been widely used for generating synthetic artworks and photographs. MidJourney is a proprietary AI image generator that specializes in producing high-quality and stylized images based on textual prompts. Unlike traditional GANs, MidJourney relies on diffusion-based generative processes and optimization techniques to create visually compelling images. DALL-E, developed by OpenAI, is another text-to-image generation model known for its ability to create creative, diverse, and highly detailed images from textual descriptions. It is particularly useful for generating conceptual art and surreal compositions. Imagen3 is an advanced AI image generation model developed by Google, emphasizing high-fidelity, photorealistic outputs. This model focuses on improved text-to-image synthesis, providing greater detail and accuracy in generating real-world and artistic images. A comparative heatmap of these models, as shown in the provided reference image, illustrates their performance across various categories, such as animals, nature, and objects.

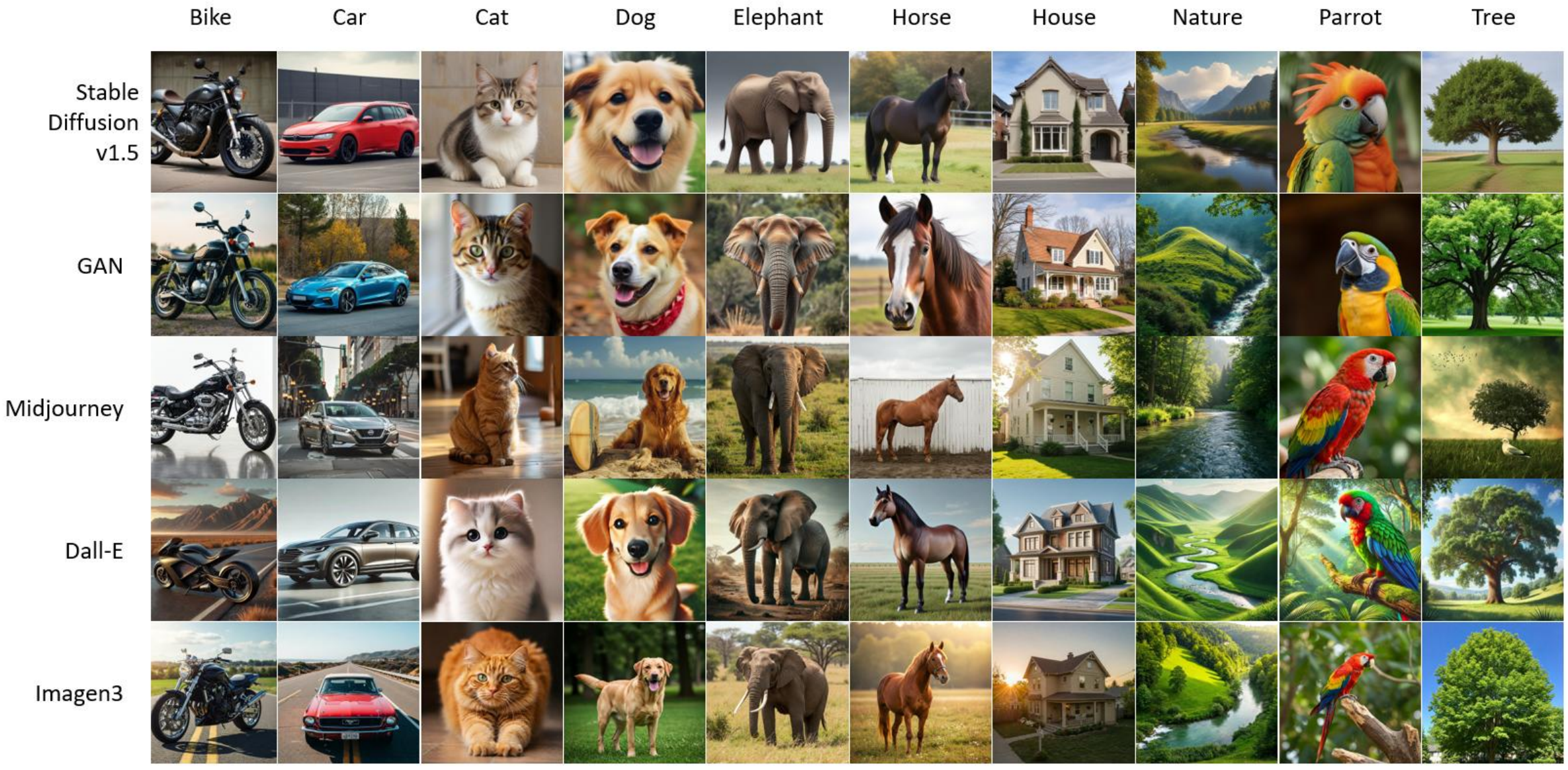

*Figure 1: Examples of Images Across Various Image Generators*

For our research, three specific image generators were selected: Stable Diffusion v1.5, GANs, and MidJourney. These models were chosen based on several criteria. The diversity in generation techniques played a crucial role in this selection. Stable Diffusion utilizes a diffusion-based approach, GANs rely on adversarial training, and MidJourney employs a refined version of text-to-image synthesis. This diversity allows for an in-depth analysis of how different generation mechanisms affect detection accuracy. Additionally, these models are among the most used AI generators in real-world applications. The selected datasets provided a balanced representation of AI-generated and real photos, ensuring class imbalance does not affect the model training, thereby enhancing the robustness of the detection approach.

The first dataset is self-generated, comprising 2000 images, with an equal distribution of 1000 real and 1000 AI-generated images in JPG format. We generated the fake image set using the Stable Diffusion v1.5 [2] model available via the ModelsLab [1] API call. The real images were sourced from various publicly available Kaggle datasets, ensuring diversity in subjects and styles.

The second dataset, Real and Fake (AI-Generated) Art Images Dataset [3], was developed in 2024 and contains 4000 images, which includes 2000 real images and 2000 AI-generated images, all in JPG format. The fake images were generated using GANs, while the real images were sourced from WikiArt, which is a comprehensive database of artworks spanning various art styles, genres, and periods.

The third dataset, MidJourney ImageNet: Real vs. Synth [7], was also released in 2024 and contains 1618 images, with 876 real and 742 AI-generated images. The real images were sourced from various parts of the internet, and the fake images were produced using MidJourney. The images across all three datasets were resized to 256x256, regardless of their original size, to ensure uniformity in model training.

## IV. PROPOSED METHODOLOGY

As mentioned prior, Generative models often struggle with accurately reproducing intricate details and complex, highly structured objects. As a result, the key to accurately distinguishing between original and synthetic images lies in the ability to identify these inconsistencies within the generated content. It is not efficient to try and detect both inconsistencies at the same time from the image. So, we introduce a two-pipeline architecture, each one specialized to detect one of the erroneous generations in the image. The first pipeline, Patchwise Feature Extraction, excels at the detection of repeat patterns of intricate details. The second pipeline – Semantic Feature Detection- is designed to detect and localize disingenuous object sampling. Before feeding the image to either pipeline, the images are resized to 256 x 256 in the RGB format.

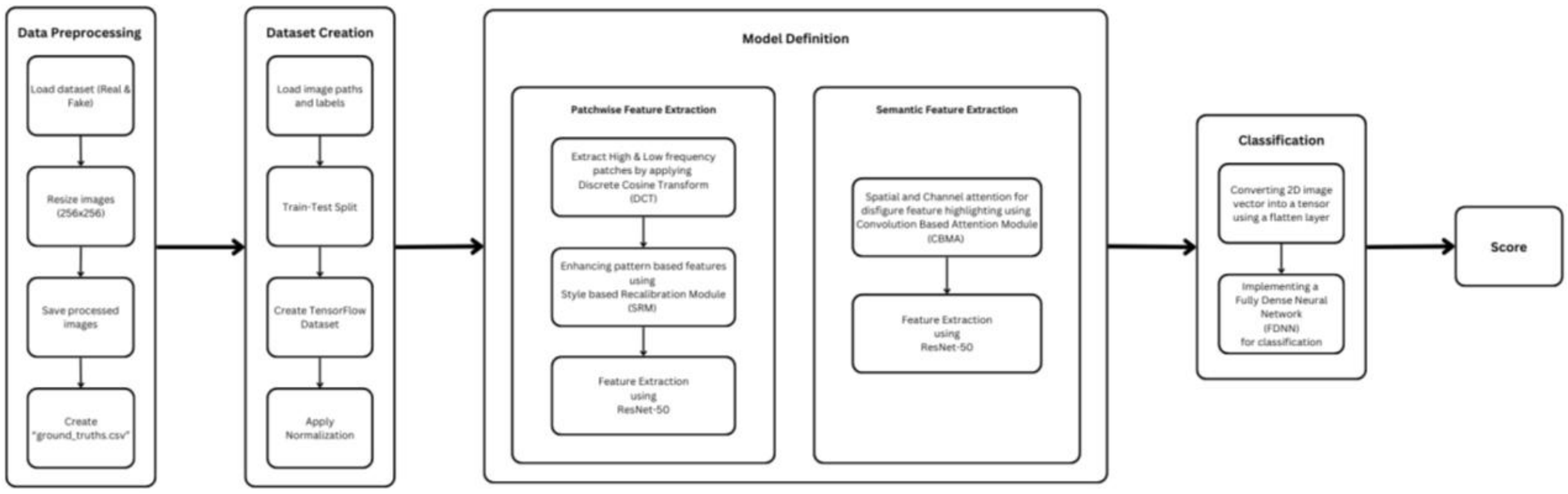

*Figure 2: Proposed Architecture Flowchart*

**Patchwise Feature Extraction:**

This pipeline consists of 3 processes, namely DCT (Discrete Cosine Transform) [17], SRM (Style Based Recalibration Module) [18] , and followed by a ResNet50 [27] . In the DCT Layer, patches of 32 x 32 are formed from the original image. A total of 64 patches are generated from every image and are each given a score based on Formula-1.

$$DCT(i,j) = \frac{1}{\sqrt{2N}} C(i)C(j) \sum_{x=0}^{N-1}\sum_{y=0}^{N-1} pixel(x,y)\, COS\left[\frac{(2x+1)i\pi}{2N}\right] COS\left[\frac{(2y+1)j\pi}{2N}\right]$$

$$C(x) = \frac{1}{\sqrt{2}}\ if\ x = 0;\ else\ 1\ if\ x > 0$$

*Equation 1: DCT*

For further processing, the two patches with the highest scores and the two patches with the lowest scores are selected. The patches with the highest scores are then concatenated channel-wise and resized to 128 x 128. The same is done with the patches with the lowest scores. Only 2 of the patches of highest and lowest are selected to have a medium for comparison while also keeping the computational overload to the minimum.

These processed patches are then sent to the SRM layer for pattern enhancement. The Style-Based Recalibration module adaptively recalibrates intermediate feature maps by exploiting their styles. SRM enhances CNNs by extracting style information from feature maps through style pooling and integrating per-channel recalibration weights. Style pooling summarizes feature responses, while style integration assigns importance to styles, adjusting feature maps to emphasize or suppress information. This returns an image of the same shape (128 x 128 x 3), which is information-heavy and ready to be trained upon by a ResNet50. The architecture for SRM can be inferred from the figure.

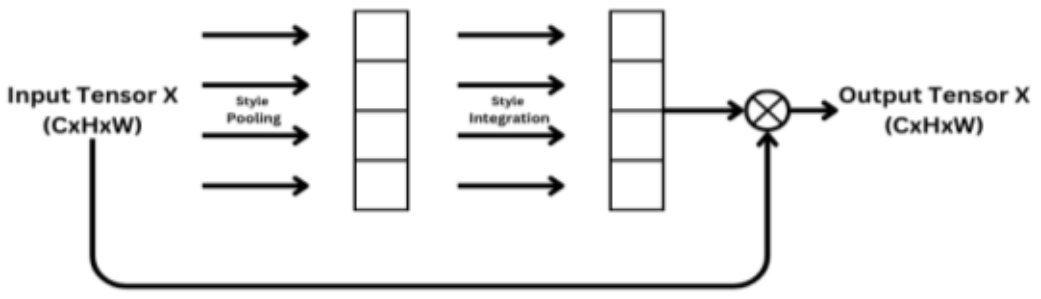

*Figure 3: Style Based Recalibration Module*

$$s = F_{srm}(X, \theta) = \sigma\Big(BN\big(CFC(SP(X))\big)\Big)$$
$$Y = sX$$

*Equation 2: SRM*

where,
$SP$ is Style Pooling,
$CFC$ is Channel wise Fully Connected,
$BN$ is Batch Normalization
$\sigma$ is the attention vector

ResNet-50 is a deep neural network that stacks residual blocks to form a 50-layer architecture. Unlike traditional networks that learn unreferenced functions, ResNets learn residual functions relative to layer inputs. Instead of forcing each layer to fit a desired mapping directly, they model the difference (residual) between the input and output, making optimization easier and improving training efficiency. The output of the ResNet50 is of shape 4 x 4 x 2048. Now, both the outputs received from the sub-pipelines of lower and higher scores means are taken. Here, the concatenation channel-wise is not taken to maintain the shape, mainly with the number of channels to give it equal weightage with the other pipeline.

**Semantic Feature Embedding:**
This pipeline consists of only two layers that efficiently highlight the disfiguration of objects generated in the image. The two key components used for this purpose are the Convolutional Block Attention Module (CBAM) [19] and ResNet50. CBAM enhances feature extraction by applying both spatial and channel attention sequentially, allowing the model to infer feature importance across these two dimensions. The CBAM architecture first refines feature representations by focusing on the most relevant spatial regions, followed by channel-wise recalibration to emphasize meaningful feature responses. The architecture for CBAM can be inferred from the figure.

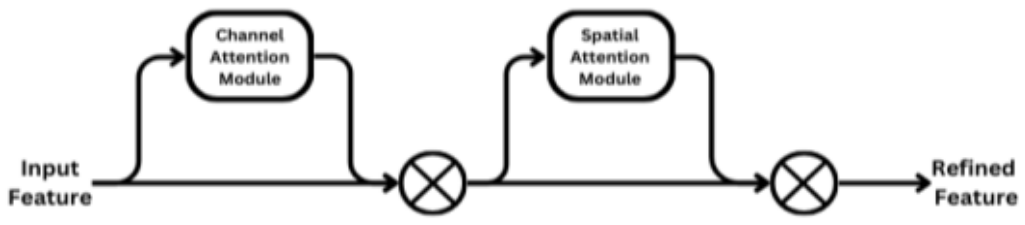

*Figure 4: Convolution Block Attention Module*

$$F' = M_c(F)F$$
$$F'' = M_c(F')F'$$

*Equation 3: CBAM*

where,
$F \in \mathbb{R}^{C\times H\times W}$ is the intermediate feature map,
$M_C \in \mathbb{R}^{C\times 1\times 1}$ is the 1D channel attention map,
$M_S \in \mathbb{R}^{1\times H\times W}$ is the 2D spatial attention map,
And $F''$ is the final refined output

Once processed through CBAM, the image is resized to 128 × 128 × 3 and passed into a ResNet50, which further extracts deep hierarchical features to distinguish between real and synthetic content effectively. Finally, an image of size 4 x 4 x 2048 matches the output size of the previous pipeline.

Using the features from both the pipelines, concatenation is done channel-wise to achieve an image shape of 4 x 4 x 4096, which is then flattened and fed into an FCNN (Fully Connected Neural Network). The FCNN has 3 layers with the 256,128, and 1 neuron at each layer. The final neuron gives us the classification as the image being real or fake, i.e., 1 if it is fake and 0 if it is real.

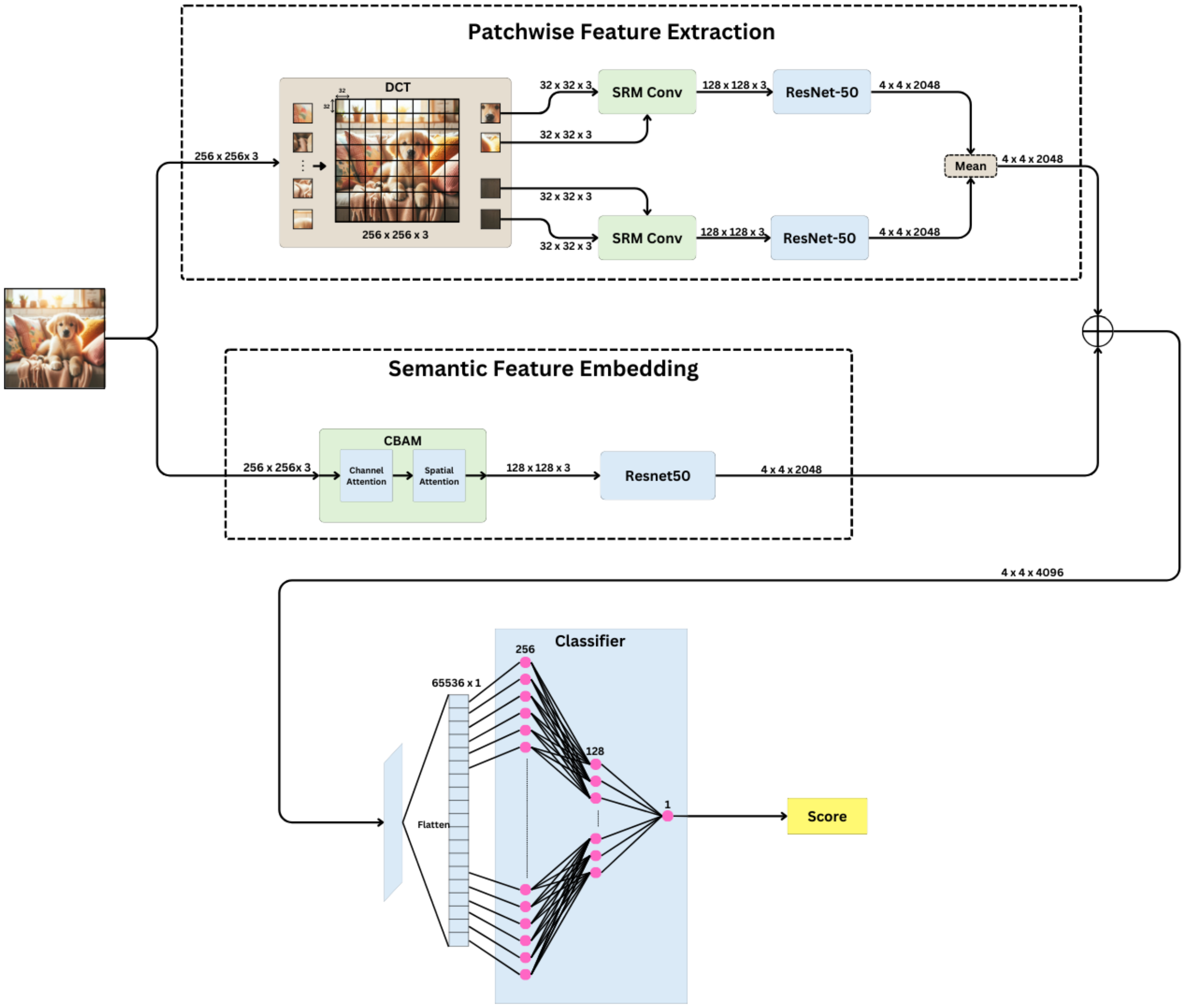

*Figure 5: Model Architecture*

**Hyperparameters:**
The training setup utilizes the Adam optimizer [20] with a learning rate of 0.0001, enabled with AMSGrad [21] and an epsilon value of 1e-6 to enhance stability.

$$m_t = \beta_1 m_{t-1} + (1-\beta_1) g_t$$
$$v_t = \beta_2 v_{t-1} + (1-\beta_2) g_t^2$$
$$\hat{v}_t = \max(\hat{v}_{t-1}, v_t)$$
$$\theta_{t+1} = \theta_t - \frac{\eta}{\sqrt{\hat{v}_t} + \epsilon} m_t$$

*Equation 4: AmsGrad*

The loss function employed is binary cross-entropy [22] as given in formula ().

$$BCE = -\frac{1}{N}\sum_{i=1}^{N}[y_i \log(p_i) + (1-y_i)\log(1-p_i)]$$

*Equation 5: Binary Crossentropy*

The model is evaluated using accuracy, precision, recall, and AUC as performance metrics. To prevent overfitting, early stopping is implemented, monitoring validation accuracy with a patience of 150 epochs in max mode. Additionally, a model checkpoint is used to save the best model based on minimum validation loss. The training process runs for a total of 200 epochs, with both early stopping and model checkpointing ensuring optimal performance.

**Metrics:**
The model was evaluated using accuracy, precision, recall, and AUC as performance metrics.

The accuracy [23] calculates the proportion of times the model succeeded in making the correct predictions against total predictions. This metric shows the general performance, but it can be deceptive in the case of imbalanced datasets because one could achieve high accuracy through a biased prediction favouring the majority class.

Precision [23] expresses the accuracy of positive predictions, showing how frequently a model has made correct positive predictions. So, high precision is desired in situations where false positives incur a high cost, e.g., in medical diagnoses.

Recall [23], on the other hand, computes the ability of the model to identify actual positive instances correctly. When there are high costs involved for false negatives, the need for high recall becomes evident. Recall expresses the capacity of a model to recognize almost all actual positive cases as such, providing highly cost-effective results in cases like fraud detection or disease diagnosis, where missing a positive case brings tremendous costs.

The AUC [24] (area under the ROC curve) describes the power of a particular model to tell positive from negative classes at various thresholds for classification. AUC is based on the plot of the true positive rate (recall) against the false positive rate. Higher values of AUC indicate better performance of the model. AUC is useful in model determination when performance assessment is required across various thresholds, particularly when doing so is a better evaluation than simple accuracy.

## V. RESULTS

The performance of our model was evaluated across three image generators: Stable Diffusion v1.5, GANs, and MidJourney. For each generator, the model's performance was assessed using various metrics, including accuracy, Area Under the Curve (AUC), loss, precision, and recall across different stack configurations, and the results are tabulated below.

*Table 1: Model Metrics*

| Image Generators | Metrics | Accuracy | AUC | Loss | Precision | Recall |
|---|---|---|---|---|---|---|
| Stable Diffusion v1.5 | Training | 1 | 1 | 5.15E-07 | 1 | 1 |
| | Validation | 0.985 | 0.989448 | 0.096024 | 1 | 0.971963 |
| GAN | Training | 1 | 1 | 8.00E-07 | 1 | 1 |
| | Validation | 0.9825 | 0.99925 | 0.057085 | 1 | 0.964286 |
| MidJourney | Training | 1 | 1 | 9.18E-07 | 1 | 1 |
| | Validation | 0.854938 | 0.864923 | 2.015429 | 0.875 | 0.87027 |

*Table 2: Metrics Comparison*

| Model | Stable Diffusion v1.5 (%) | GAN (%) | MidJourney (%) | Average (%) |
|---|---|---|---|---|
| ResNet-50 [27] | 99.70 | 52.00 | 54.90 | 68.86 |
| CNNSpot [28] | 95.90 | 46.80 | 52.80 | 65.16 |
| F3Net [29] | **99.90** | 49.90 | 50.10 | 66.63 |
| GramNet [30] | 99.10 | 51.70 | 54.20 | 68.33 |
| PatchCraft[10] | 89.30 | 72.40 | 79.00 | 80.23 |
| UnivFD [14] | 84.00 | 80.30 | 73.20 | 79.16 |
| Shilin Yan et al's Model [26] | 99.7 | 72.40 | 79.38 | 83.82 |
| Our Model | 98.5 | **98.25** | **85.49** | **94.08** |

For the Stable Diffusion v1.5 image generator, the model achieved an accuracy of 98.5%, with a corresponding loss of 0.096. The precision and recall were recorded at 100% and 97.19%, respectively, while the AUC was measured at 0.989.

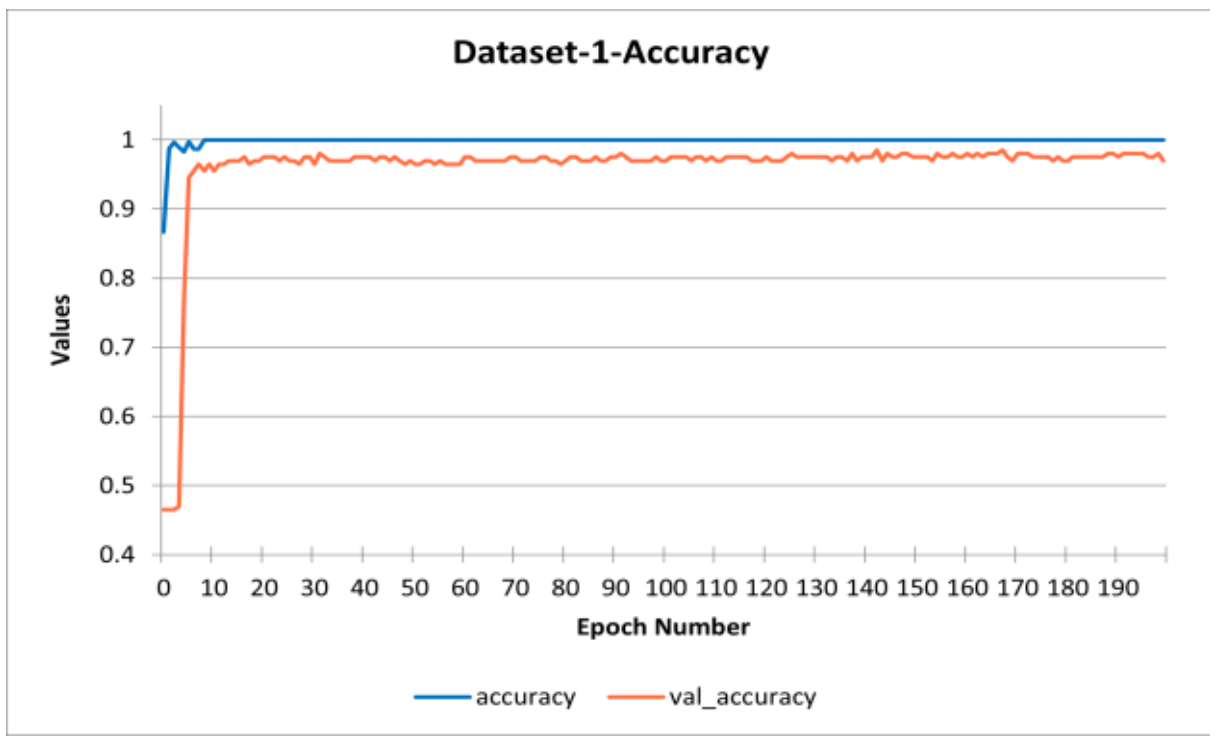

*Figure 6: Stable Diffusion-Accuracy*

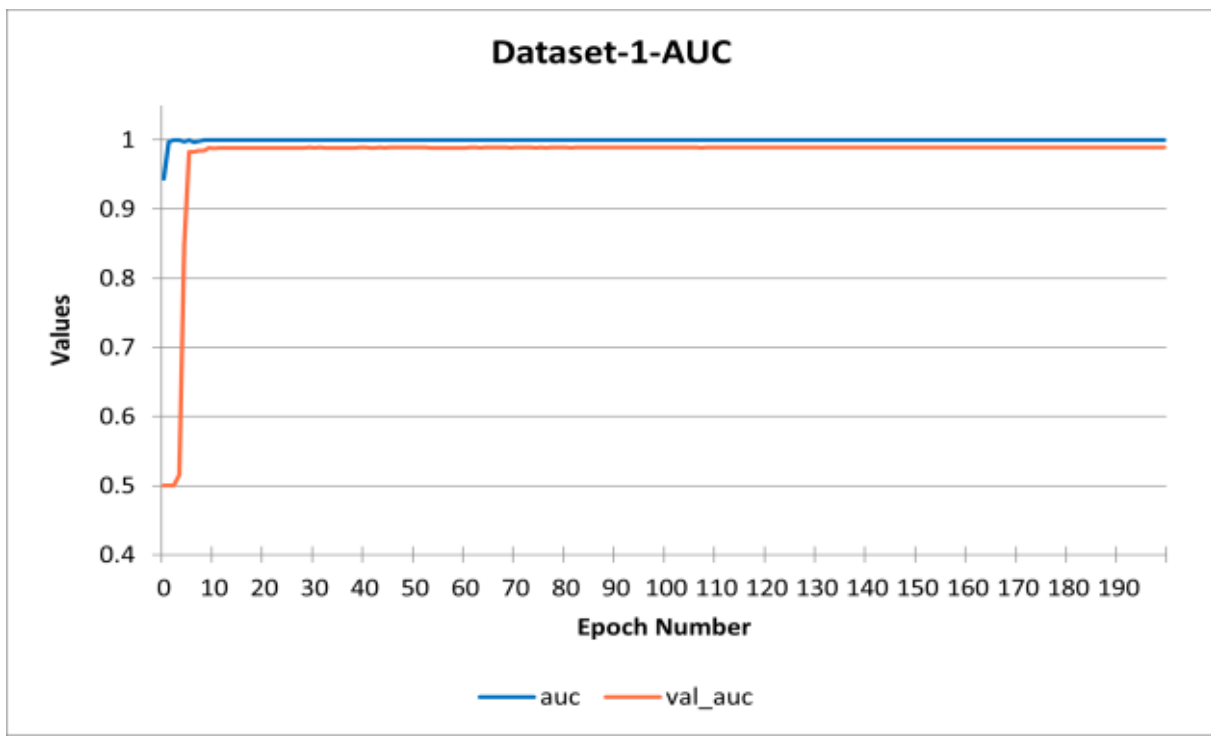

*Figure 7: Stable Diffusion-AUC*

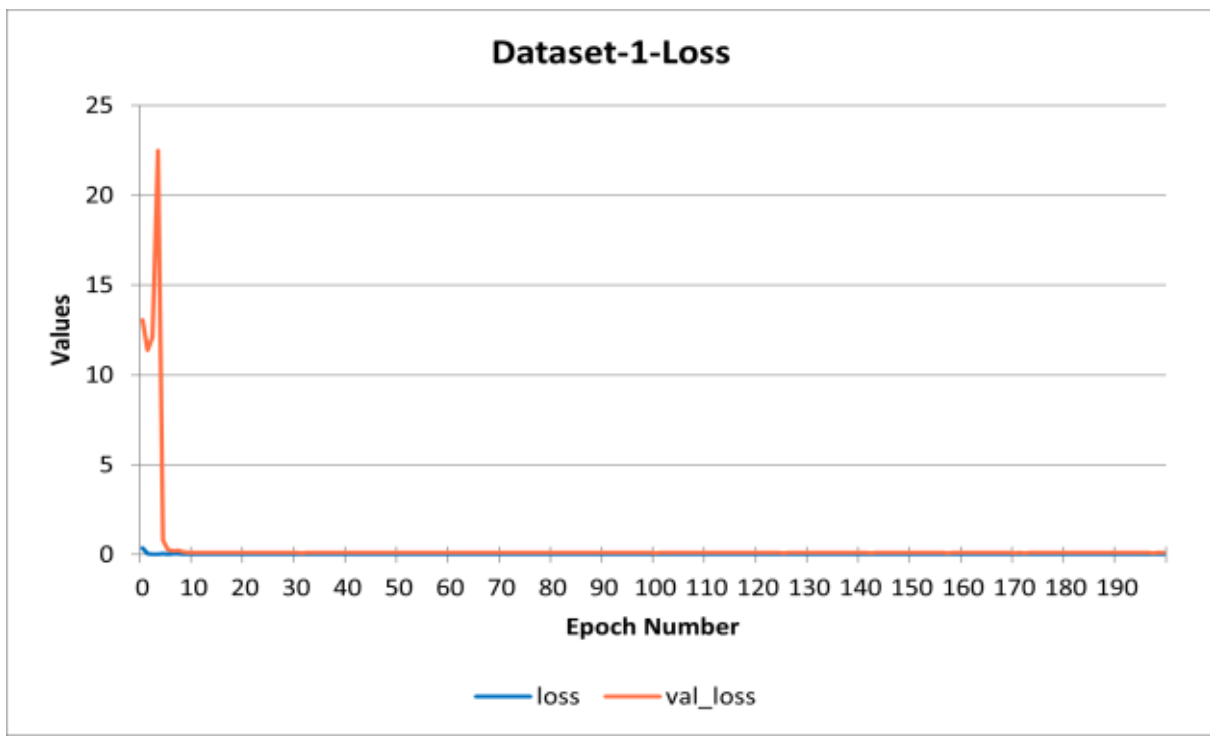

*Figure 8: Stable Diffusion-Loss*

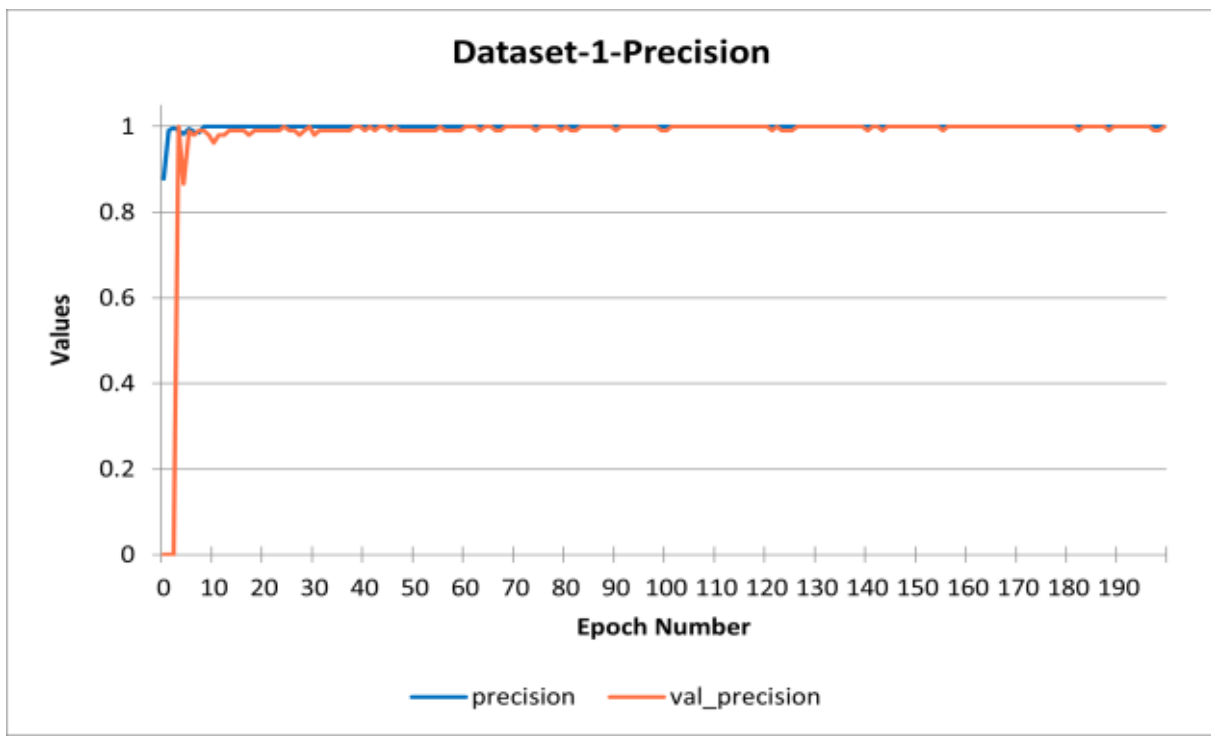

*Figure 9: Stable Diffusion-Precision*

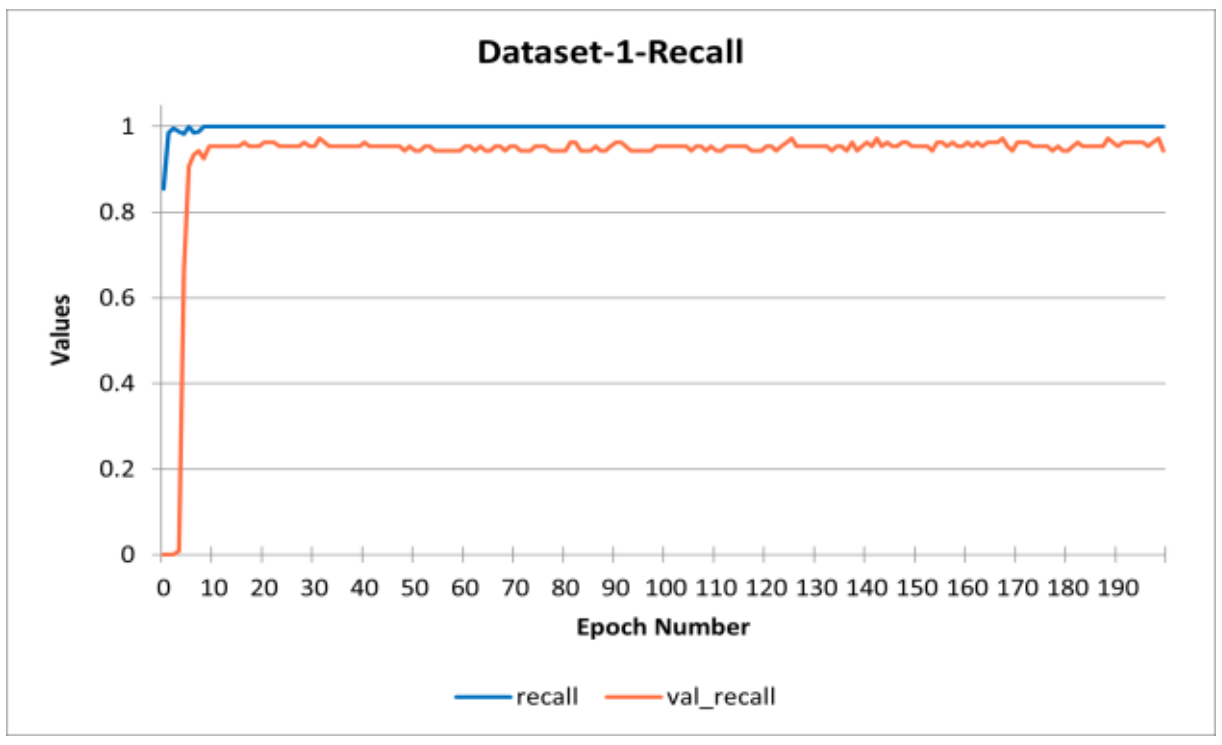

*Figure 10: Stable Diffusion-Recall*

For the GAN image generator, the model exhibited an accuracy of 98.25%, with a loss of 0.057. The precision and recall values stood at 100% and 96.4%, respectively, while the AUC was 0.99.

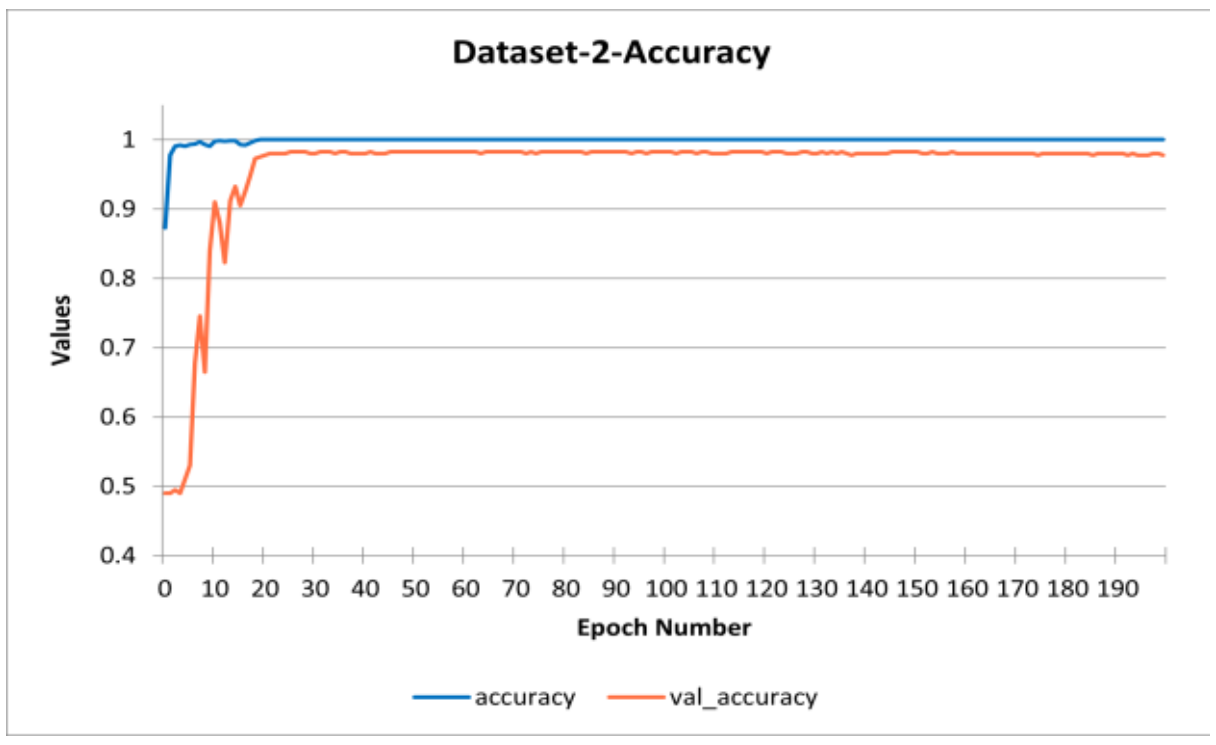

*Figure 11: GAN-Accuracy*

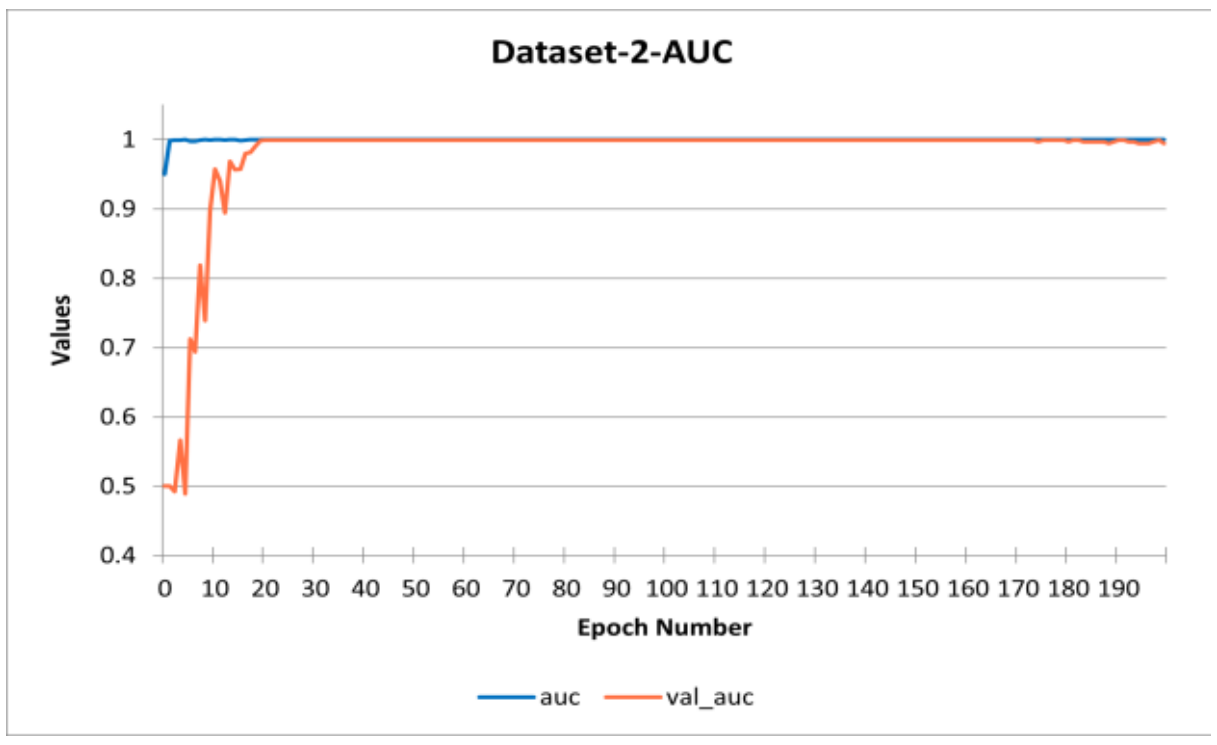

*Figure 12: GAN-AUC*

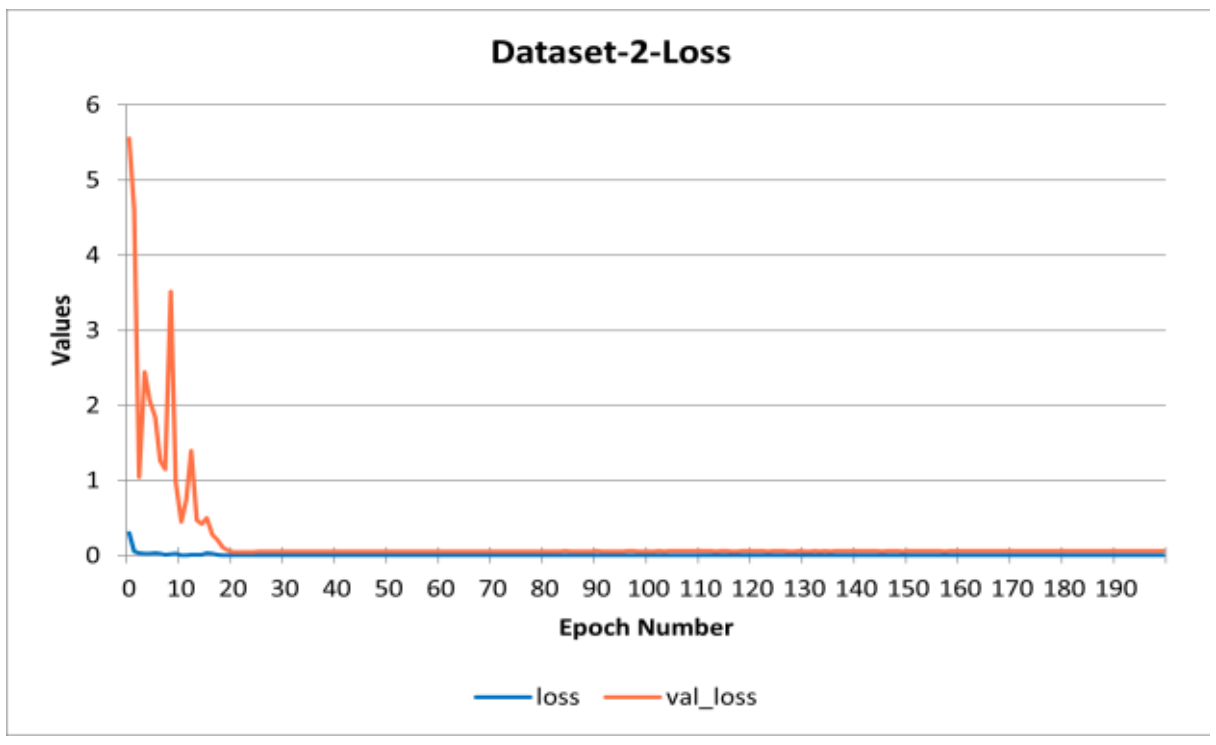

*Figure 13: GAN-Loss*

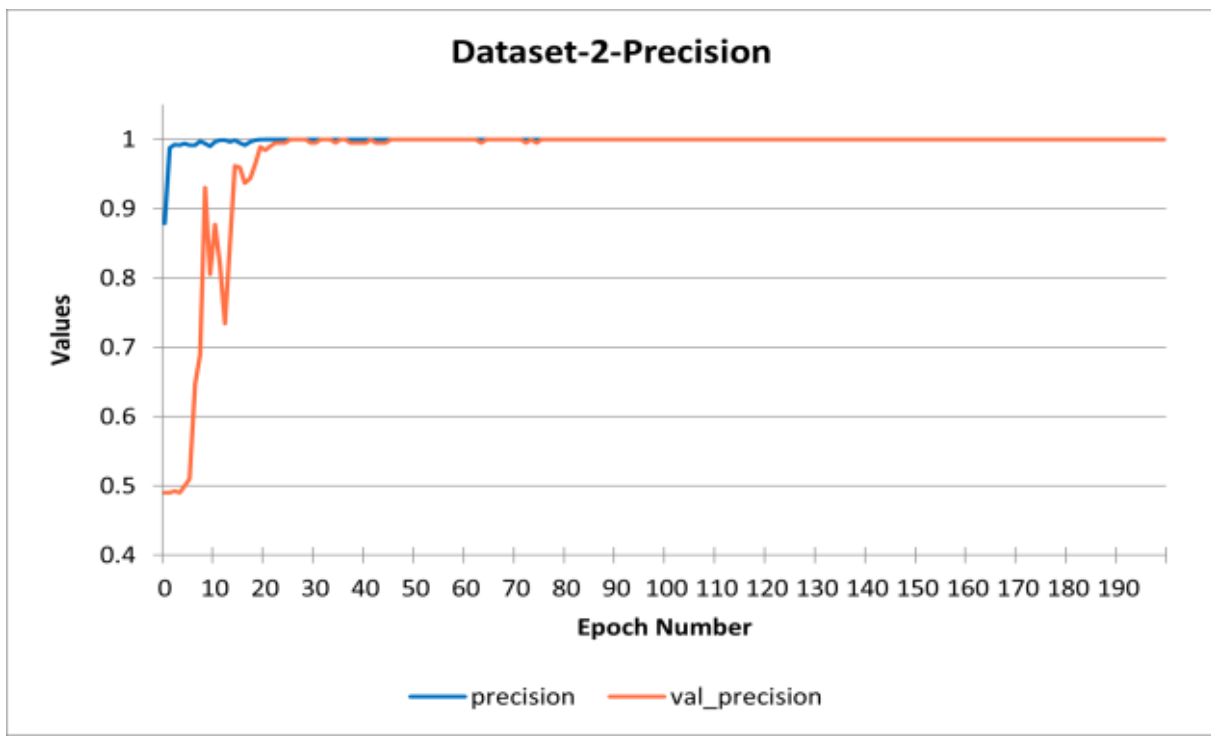

*Figure 14: GAN-Precision*

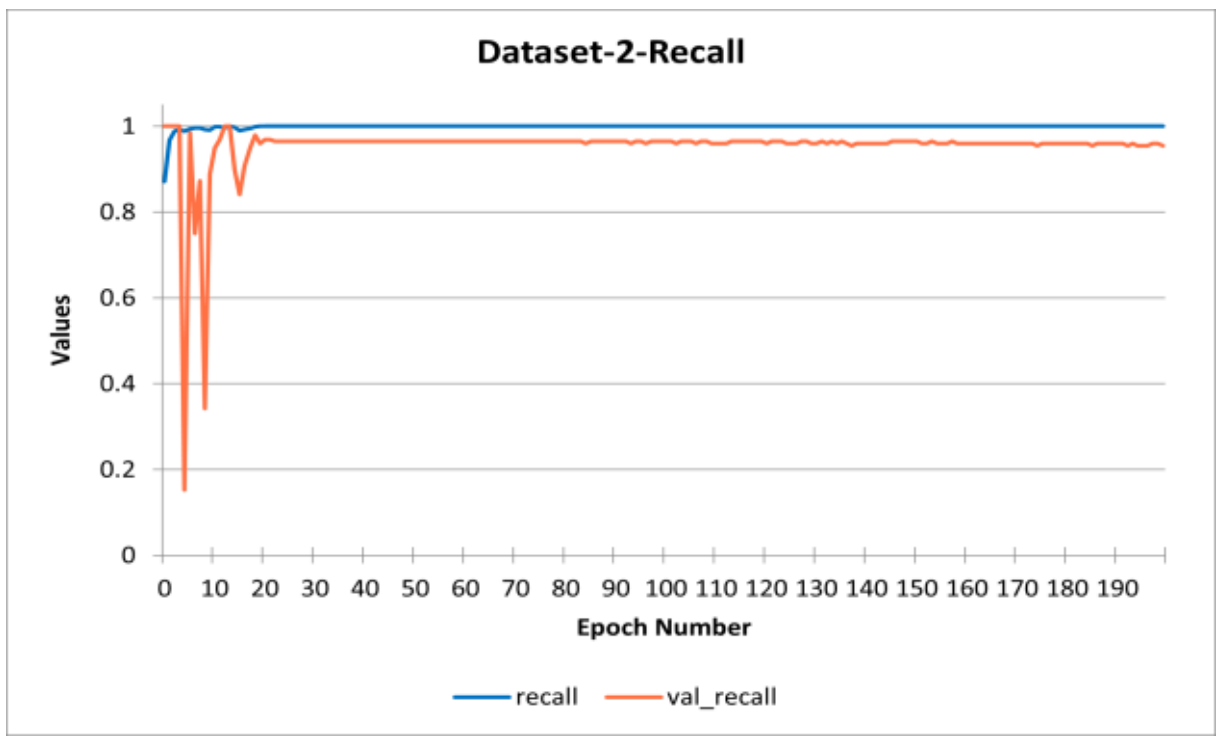

*Figure 15: GAN-Recall*

For the MidJourney dataset, the model produced an accuracy of 85.49%, with a loss of 2.015. The precision and recall were found to be 87.5 and 87.02, respectively, with an AUC of 0.87.

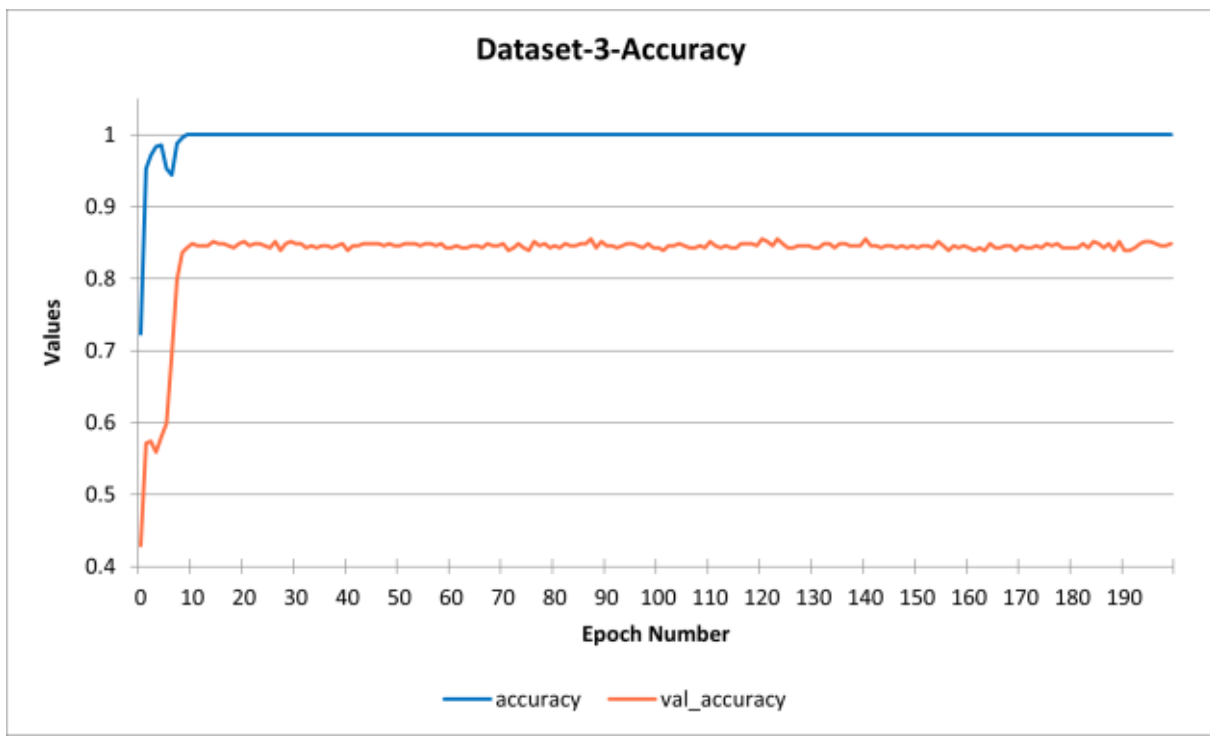

*Figure 16: Midjourney-Accuracy*

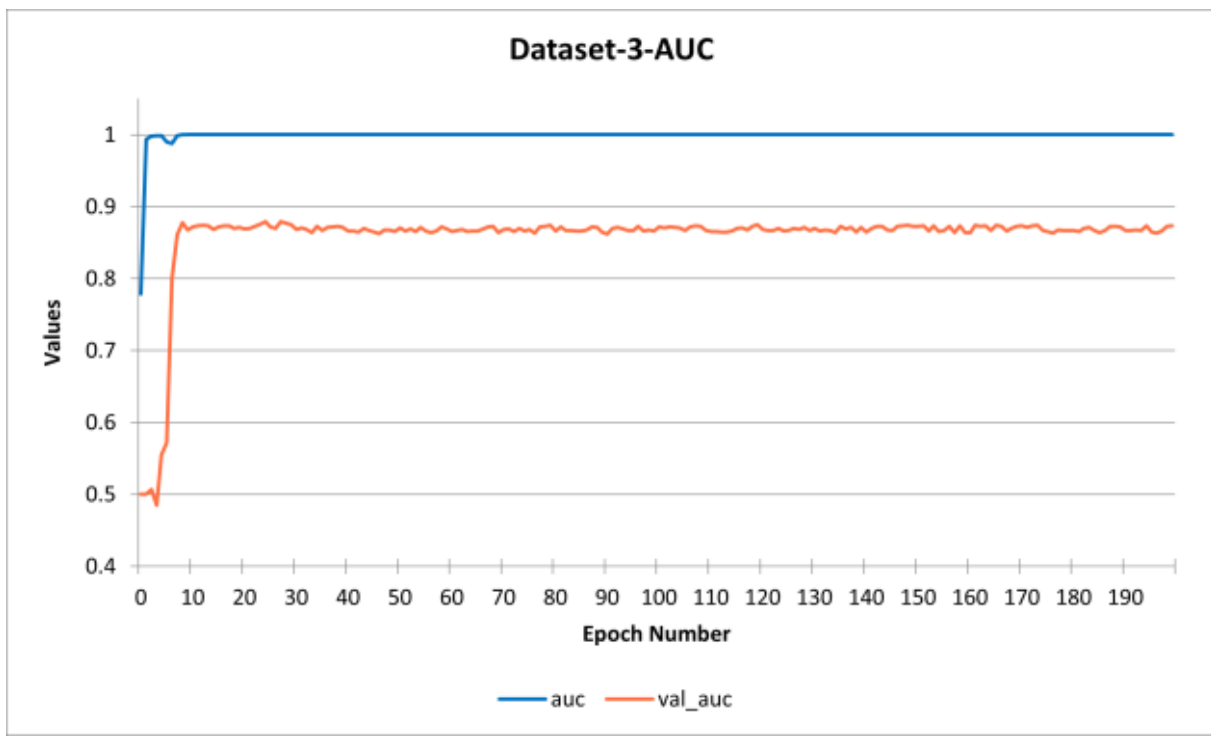

*Figure 17: Midjourney-AUC*

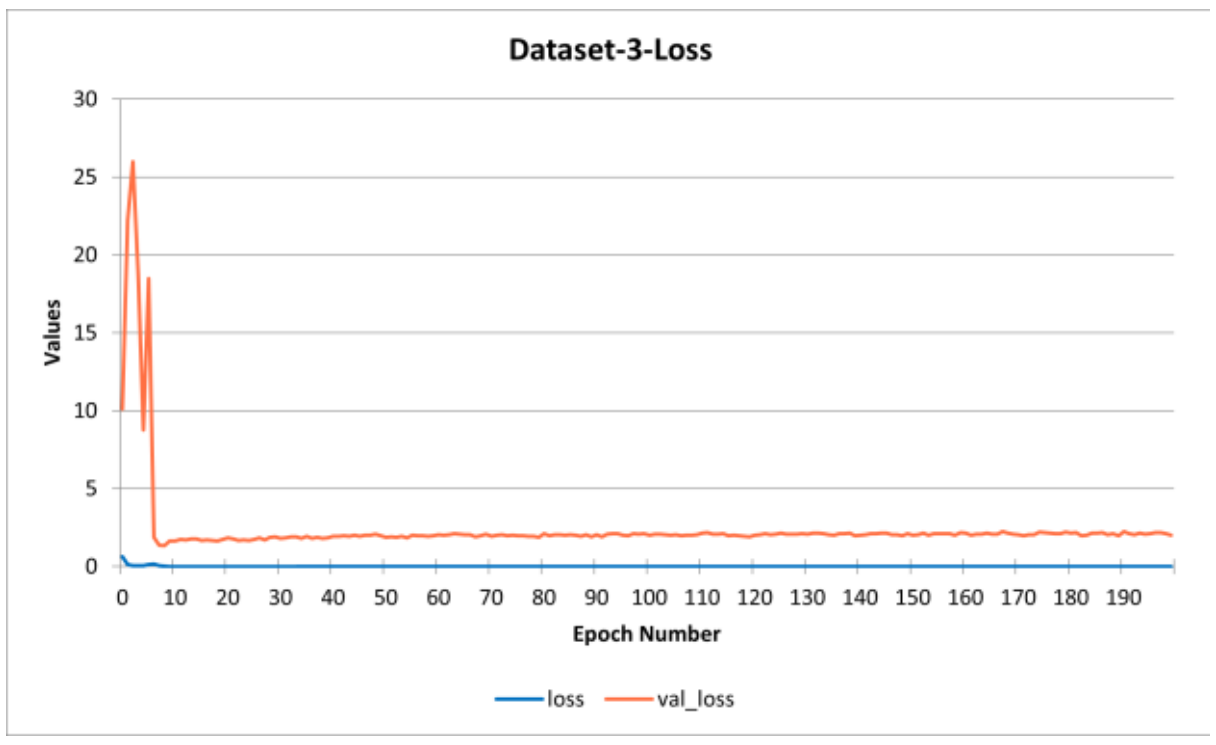

*Figure 18: Midjourney-Loss*

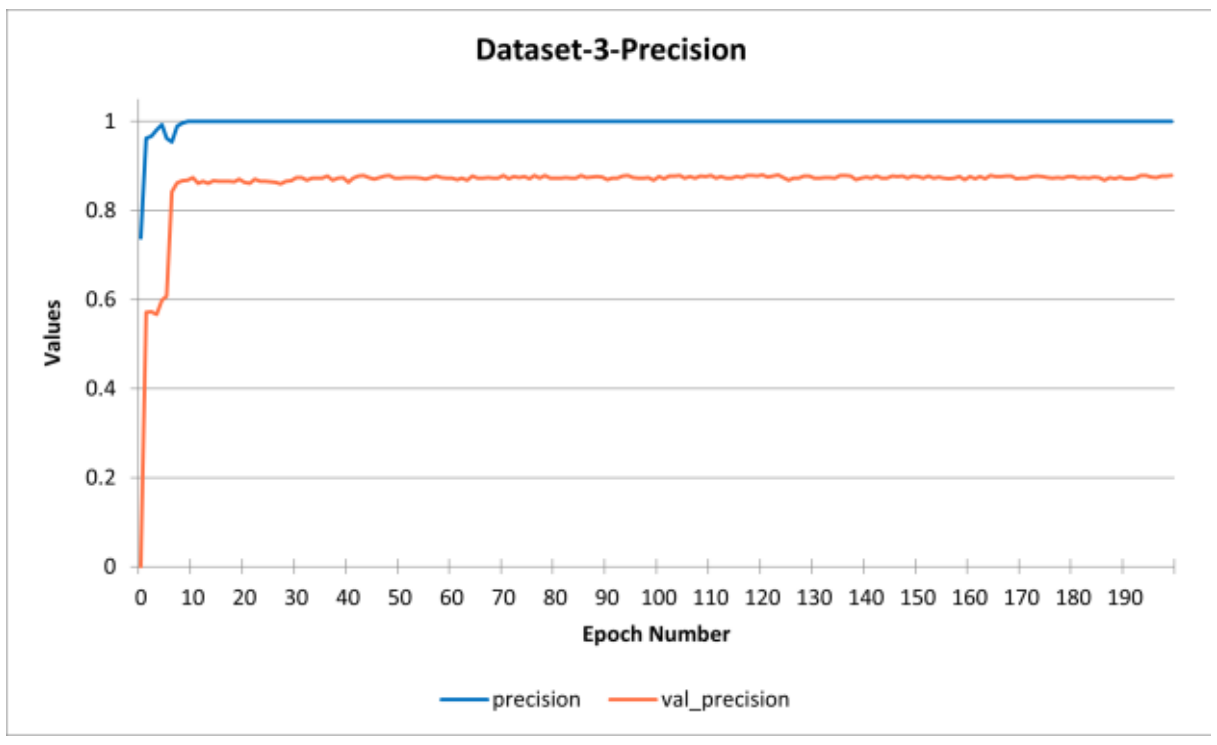

*Figure 19: Midjourney-Precision*

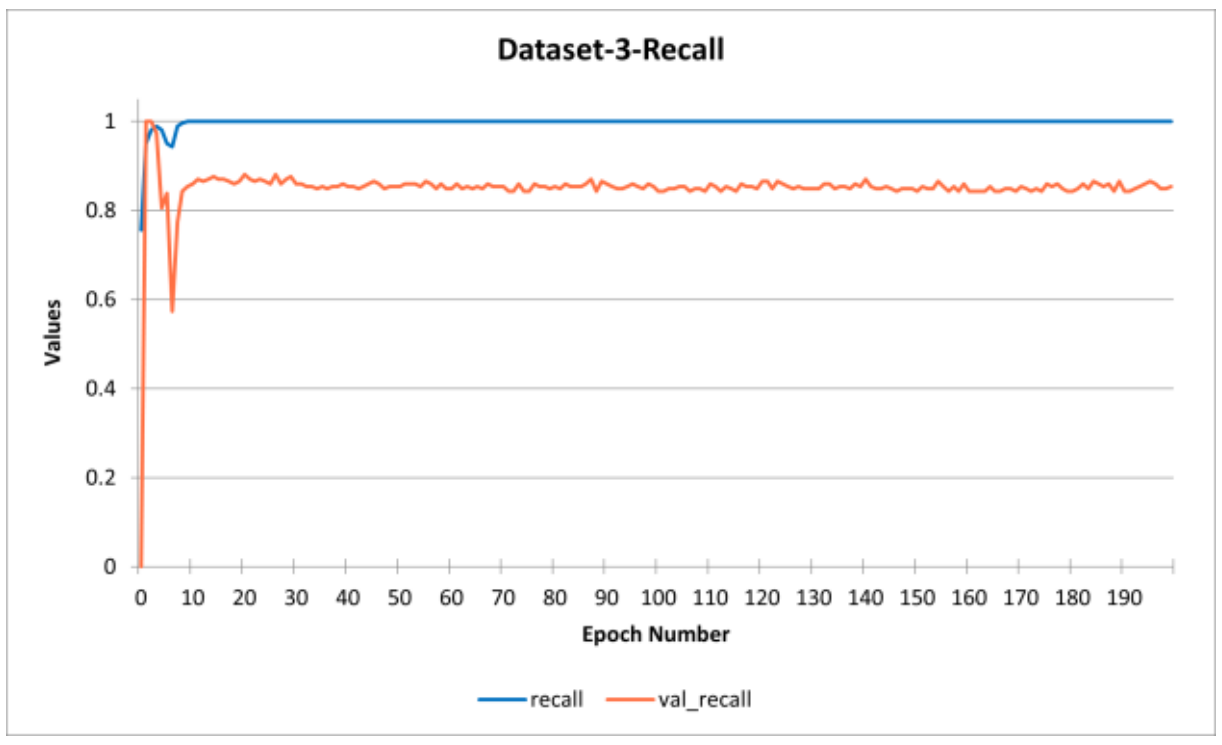

*Figure 20: Midjourney-Recall*

A comparative analysis of the three datasets highlights that two datasets demonstrated superior classification accuracy, while one posed greater challenges. The differences in performance across datasets suggest that certain types of AI-generated images are more difficult to classify, possibly due to the sophistication of the generation techniques or dataset-specific artifacts. Understanding these variations is crucial for improving model robustness and generalizability.

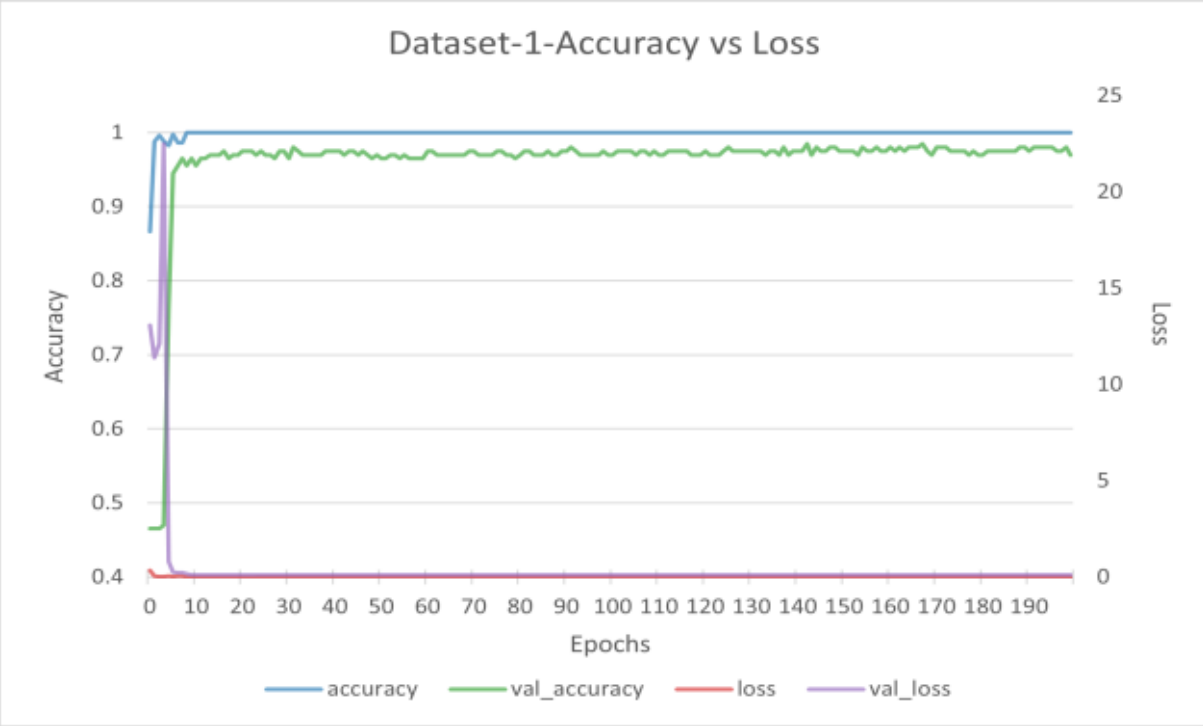

*Figure 21: Stable Diffusion-Accuracy vs Loss*

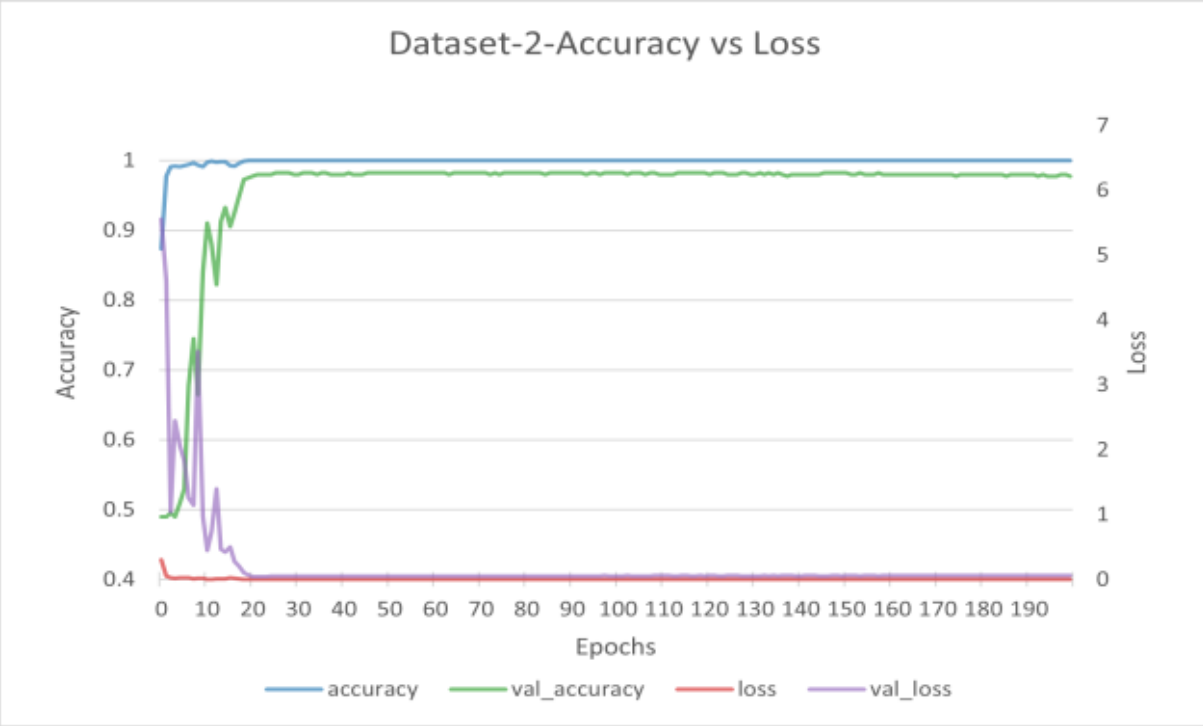

*Figure 22: GAN-Accuracy vs Loss*

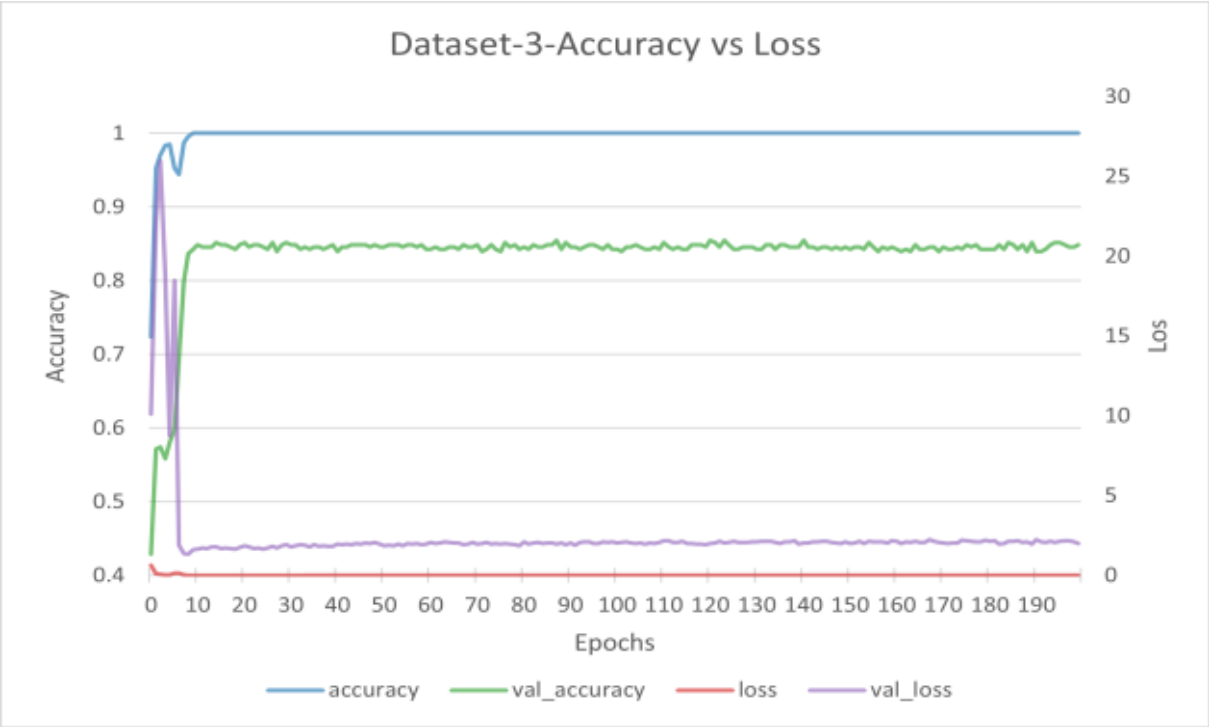

*Figure 23: Midjourney-Accuracy vs Loss*

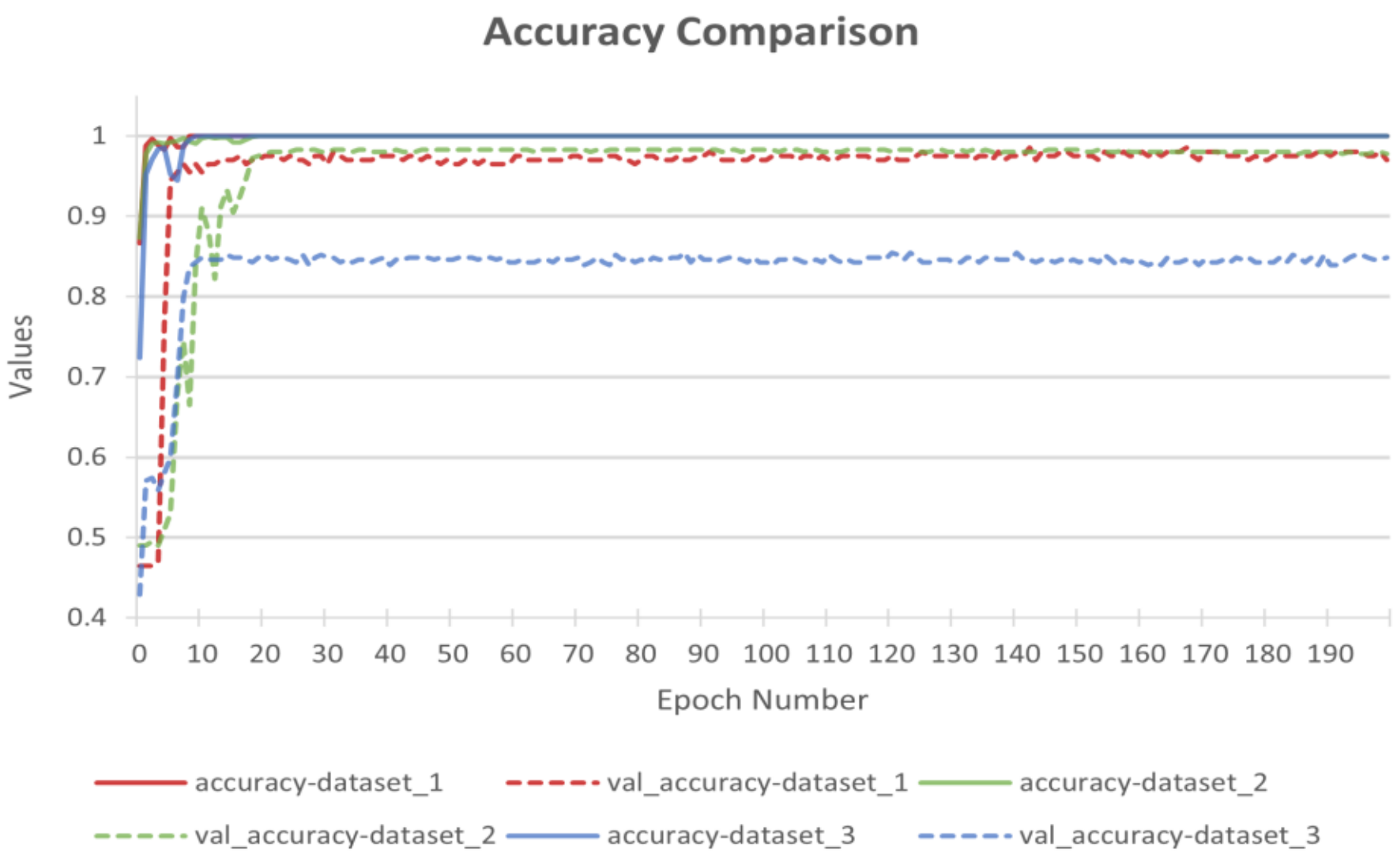

*Figure 24: Accuracy Comparison*

## VI. CONCLUSION

This paper demonstrates the effectiveness of our custom deep learning architecture in distinguishing AI-generated images from real ones. By leveraging a combination of patch-wise feature extraction, semantic feature embedding, and a robust classification module, we achieved high accuracy across multiple datasets, including ModelsLab, GAN, and Stable Diffusion v1.5. Our results highlight the impact of combining DCT, SRM Conv, and ResNet-50 for detailed feature extraction, while CBAM-enhanced ResNet-50 further refines the semantic understanding of images.

Across all datasets, our model consistently delivered strong performance, achieving high AUC and precision and showcasing its reliability in detecting AI-generated content. Integrating multiple feature extraction techniques proved beneficial in capturing both low-level statistical patterns and high-level semantic differences between real and generated images.

These findings reinforce the importance of hybrid feature extraction strategies in deepfake detection, offering a scalable and effective approach for real-world applications.

Future work could explore expanding the model to handle more advanced generative models, optimizing computational efficiency, and integrating adversarial training to enhance robustness against evolving AI-generation techniques. Additionally, the inclusion of larger and more diverse datasets could further improve generalization and real-world applicability.